\documentclass[default]{sn-jnl}% Default
\usepackage{amssymb}
\usepackage{lipsum}
\usepackage{graphicx}
\usepackage{subcaption}
\usepackage{adjustbox}
\usepackage{booktabs}
\usepackage{ragged2e}
\usepackage{array}
\usepackage{amsmath}
\usepackage{siunitx} 
\usepackage{adjustbox}
\usepackage{graphicx}%
\usepackage{multirow}%
\usepackage{amsmath,amssymb,amsfonts}%
\usepackage{amsthm}%
\usepackage{mathrsfs}%
\usepackage[title]{appendix}%
\usepackage{xcolor}%
\usepackage{textcomp}%
\usepackage{manyfoot}%
\usepackage{booktabs}%
\usepackage{algorithm}%
\usepackage{algorithmicx}%
\usepackage{algpseudocode}%
\usepackage{listings}%
\theoremstyle{thmstyleone}%
\theoremstyle{thmstyletwo}%

\theoremstyle{thmstylethree}%

\begin{document}

\title[Article Title]{DynaHull: Density-centric Dynamic Point Filtering in Point Clouds}


\author*[1,2]{\fnm{Pejman} \sur{Habibiroudkenar}}\email{pejman.habibiroudkenar@aalto.fi}

\author[1]{\fnm{Risto} \sur{Ojala}}\email{risto.j.ojala@aalto.fi}
%%\equalcont{These authors contributed equally to this work.}

\author[1,2]{\fnm{Kari} \sur{Tammi}}\email{kari.tammi@aalto.fi}
%%\equalcont{These authors contributed equally to this work.}

\affil[1]{\orgdiv{Mechanical Engineering}, \orgname{Aalto University}, \orgaddress{\city{Espoo} \postcode{02150} \country{Finland}}}
\affil[2]{\orgdiv{Helsinki Institute of Physics} \orgname{University of Helsinki}, \orgaddress{\city{Helsinki} \postcode{00014}  \country{Finland}}}
%\affil*[3]{\orgdiv{Mechanical Engineering}, \orgname{Aalto University}, \orgaddress{\street{}, \city{Espoo}, \postcode{02160}, \state{Uusima}, \country{Finland}}}

\abstract{In the field of indoor robotics, accurately localizing and mapping in dynamic environments using point clouds can be a challenging task due to the presence of dynamic points. These dynamic points are often represented by people in indoor environments, but in industrial settings with moving machinery, there can be various types of dynamic points. This study introduces DynaHull, a novel technique designed to enhance indoor mapping accuracy by effectively removing dynamic points from point clouds. DynaHull works by leveraging the observation that, over multiple scans, stationary points have a higher density compared to dynamic ones. Furthermore, DynaHull addresses mapping challenges related to unevenly distributed points by clustering the map into smaller sections. In each section, the density factor of each point is determined by dividing the number of neighbors by the volume these neighboring points occupy using a convex hull method. The algorithm removes the dynamic points using an adaptive threshold based on the point count of each cluster, thus reducing the false positives. The performance of DynaHull was compared to state-of-the-art techniques, such as ERASOR, Removert, OctoMap + SOR , and Dynablox, by comparing each method to the ground truth map created during a low activity period in which only a few dynamic points were present. The results indicated that DynaHull outperformed these techniques in various metrics, noticeably in the Earth Mover’s Distance, false negatives and false positives.}

\keywords{Dynamic Points Removal, Point Cloud, SLAM, ConvexHull}



\maketitle

\section{Introduction}\label{sec1}

Autonomous robotics have become increasingly popular in recent years in research in which indoor positioning and global navigation systems have been of particular interest \cite{LOGANATHAN2023101343}. Mapping stands out as a crucial element in the indoor robotics ecosystem, being essential for several tasks, such as localization, robot control, route-planning, perception, and overall system management. Localization in outdoor settings is often supported by GPS systems, but this traditional method lacks accuracy for indoor environments due to the limitations of the line-of-sight (LoS) path from the satellite \cite{app11010279}. Moreover, indoor mapping primarily relies on RGB cameras \cite{robotics11010024} and 2D LiDAR \cite{2dlidar}, with fewer studies exploring the utilization of 3D LiDAR for indoor mapping as 3D LiDAR has been mainly used in outdoor environments \cite{outdoor3d}. To establish an accurate map, the robot must achieve precise localization; conversely, effective localization is contingent upon a clean, stable map. This interdependence is addressed through the implementation of a technique known as Simultaneous Localization and Mapping (SLAM), a concept that has been set in motion over the past three decades \cite{slam}. Mapping in highly dynamic environments, such as in a warehouse teeming with moving people and operational automation robots, presents a unique challenge.  The mapping process in such settings is prone to capturing ’dynamic points’, also known as ghost or phantom points, which can compromise the accuracy of a robot’s scan matching localization in subsequent operations due to the present noise. Removing these phantom points, not only significantly enhances localization accuracy and reduces drift \cite{LOAMsuma}, but can also enable creation of a clean and accurate map capable of being used for various reconstruction purposes. For instance, it can facilitate the generation of a digital representation of the environment; this reconstructed map would provide valuable insights and can be leveraged in multiple applications. In the context of industrial settings, the reconstructed map can be used to optimize material flow and analyze the efficiency of construction operations. The proposed dynamic point removal method in this study relies on the distinctive difference in density between the stationary and dynamic points. After several scans, the stationary points are close to each other while the dynamic points are sparse due to their movement. Our algorithm identifies dynamic points based on the sparsity of their neighboring points. To alleviate the non-uniformly distributed points in mapping processes, the filtering algorithm proposed in this study also clusters the map into smaller sections with different densities. 
\subsection{Contributions}

This study introduces a novel post-mapping, density-based dynamic point removal technique called DynaHull. DynaHull leverages a convex hull to handle point clouds with non-uniformly distributed points. Additionally, this research customizes a dynamic point removal process specifically for indoor environments, diverging from conventional methods used in outdoor mapping. Unlike state-of-the-art techniques that require two or more consecutive scans to detect and remove dynamic points in a point cloud, our method needs only the final map without prior scans. Our approach outperformed state-of-the-art dynamic point removal methods in a highly dynamic indoor environment.

\section{Related Work}\label{sec2}

Dynamic points in point clouds resulting from dynamic objects undermine LiDAR scan matching reliability resulting in lower localization accuracy. In response, researchers have developed several strategic approaches, each one addressing dynamic point issue from a different technological and methodological standpoint. In broad term,  dynamic points removal in a point cloud can be divided into two groups, Geometrical and Machine learning based. Machine learning-based methods, as referenced in \cite{rangenet,atg,voxelnet,PointRCNN}, involve segmenting environmental data into distinct entities. This segmentation helps the system recognize entities producing 'ghost tails,' such as people, cars, and bicycles, which are subsequently removed from the point cloud. Notably, Ayush Dewan and colleagues \cite{atg} devised a method using motion cues for detecting and tracking dynamic objects in urban environments, bypassing the need for prior maps. They employed Bayesian methods for segmentation and RANSAC for motion model estimation. Complementarily, Zhou et al.’s VoxelNet  \cite{voxelnet} bypassed traditional hand-crafted feature representations, using an end-to-end trainable network for directly processing sparse 3D points. This has significantly improved LiDAR-based detection of cars, pedestrians, and cyclists. Furthermore, \cite{PointRCNN} introduced PointRCNN, a two-stage framework for 3D object detection from point clouds, which has advanced proposal generation and refinement techniques. Despite these advancements, segmentation-based strategies have limitations. A notable challenge is their struggle with unknown classes \cite{unknownclass} and incomplete detection in cases of occlusion. This highlights the ongoing need for developing more robust, versatile, and accurate dynamic point segmentation methods in LiDAR data-processing.

Although machine learning techniques offer significant advantages, this paper primarily focuses on geometrical dynamic point removal, as this topic necessitates dedicated research. These solutions mainly include ray-tracing methods that consider space occupancy, visibility-based methods focusing on logical geometric consistencies, segmentation methods using distinctive features of stationary and dynamic points. Furthermore, the techniques described aim to eliminate dynamic points at two distinct stages of execution: the first group does so by removing dynamic points with each scan, termed 'real-time' in this study; the second group removes dynamic points after the mapping process, which is referred to as 'post-mapping'.

\subsection{Voxel Ray Casting Methods}

Voxel ray-casting methods, commonly known as ’ray-tracing,’ effectively manage the removal of dynamic objects by adjusting the likelihood of space being occupied along the path of the ray as referenced in various studies \cite{ChangeDetection,Robust,Peopleremover,octomap,dynamicbench}. Areas previously occupied by moving objects become increasingly likely to be considered free as more rays intersect with that space. The early work in this domain, grounded in Dempster-Shafer Theory (DST), also known as the theory of belief, utilizes probability equations to determine the occupation status of a voxel. Xiao et al. \cite{ChangeDetection} enhanced the reliability of spatial representations by fusing occupancy information and integrating data from adjacent rays using the Weighted Dempster-Shafer theory. Building on this, Georgi Postica et al. \cite{Robust} further refined dynamic points detection by discretizing the occupancy representation based on its distance from the sample scan origin in DST, thus significantly reducing false positives and increasing both speed and accuracy. Schauer et al. \cite{Peopleremover} improved this approach by traversing a voxel occupancy grid along the lines of sight between the sensor to find differences in volumetric occupancy between the scans, resulting in an increase of the number of scans used for calculations and reducing the required parameter tuning to just one: voxel size. To address the computational cost challenges associated with voxel-based methods, the OctoMap  \cite{octomap} method utilized an efficient oct-tree data structure. Building on this, Kin-Zhang et al. \cite{dynamicbench} further optimized the approach by minimizing the impact of noise and abnormal points in the OctoMap using the Statistical Outlier Removal (SOR) technique for filtering, and Sample Consensus for performing ground segmentation. In this study this method is refereed to as OctoMap + SOR. Despite their precision and utility, ray-tracing methods face the challenge of incidence angle ambiguity. As the range of measurement extends, the ambiguity in the incidence angle of a point increases, leading to the potential misidentification of adjacent ground points as dynamic objects. This limitation has spurred the development of visibility-based techniques.

\subsection{Visibility} 

\begin{figure}[h]
    \centering
    \includegraphics[width=0.99\linewidth]{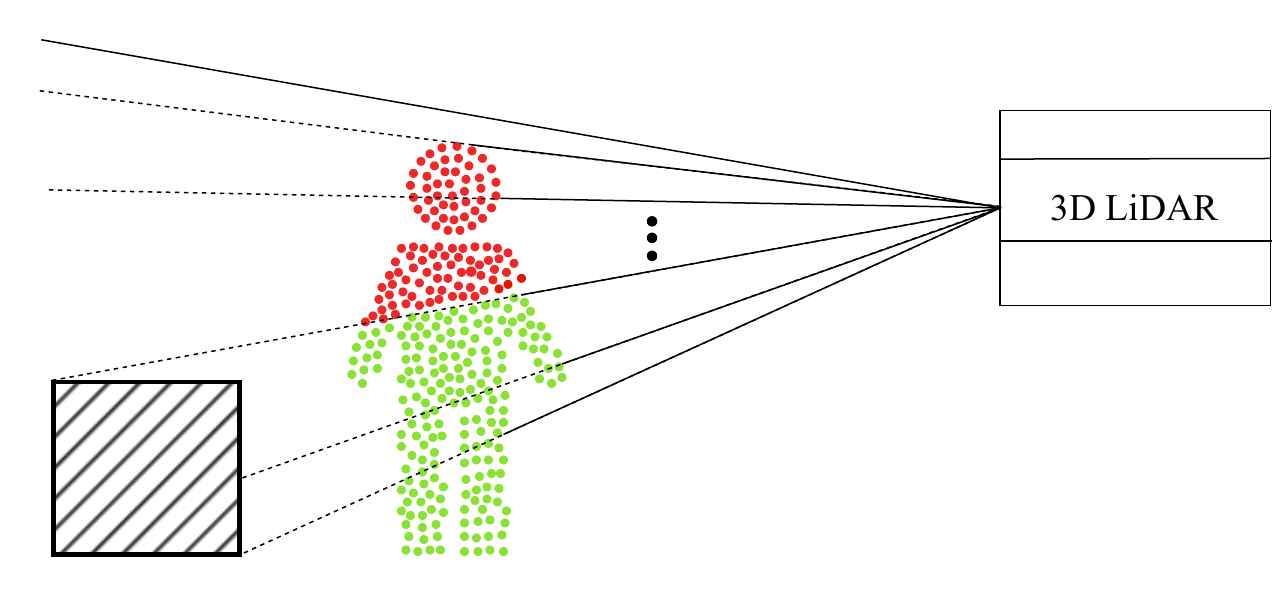}
    \caption{Irremovable dynamic points are presented in the color red because there are no stationary objects present behind those points. Meanwhile, the green areas are removed because there is a stationary box behind the observed point.}
    \label{fig:idp}
\end{figure}

Visibility-based methods  \cite{pomerleau2014,metarooms,randr2020} operate under the assumption that if a query point is discerned behind a point previously integrated into the map, it indicates that the initial point is dynamic. These approaches require finding an association between a single point and a map point. Consequently, \cite{pomerleau2014,metarooms} have focused on iteratively updating the dynamic and stationary states by repeatedly fusing multiple measurements and checking the visibility between a query scan and a map. However, pose-estimate errors accumulated in LiDAR motion can render this association inaccurate, thus raising the possibility of erroneously deleting stationary points. To address this, Kim \textbf{Removert} \cite{randr2020} proposed a multiresolution range image-based false prediction-reverting algorithm, also known as Removert. Initially, this method retains certain static points, then progressively reinstates more ambiguous static points by enlarging the query-to-map association window. This approach compensates for errors in LiDAR motion or registration, hence the name ”remove, then revert.” However, visibility-based methods struggle with occlusion and the elimination of dynamic points at which static points were not detectable behind these dynamic elements commonly known as Irremovable dynamic point, as demonstrated in Fig. \ref{fig:idp}. This challenge impedes accurately differentiating between static and dynamic points, which is crucial for creating reliable and detailed maps. To address these challenges and reduce false positives, segmentation methods have gained popularity.

\begin{table*}[h]
    \caption{Dynamic point removal methods and their limitations}
    \centering
    \renewcommand{\arraystretch}{1}
    \setlength{\tabcolsep}{5pt}
    \begin{adjustbox}{width=1\textwidth}
    \fontsize{8}{8}\selectfont
        \begin{tabular}{@{}>{\centering\arraybackslash}p{3cm}p{6cm}>{\centering\arraybackslash}p{2cm}>{\centering\arraybackslash}p{2cm}p{2cm}p{4cm}@{}}
            \toprule
            \textbf{Author} & \textbf{Paper} & \textbf{Method Type} &  \textbf{Designed for indoors} &  \textbf{Removal time}\\\\
            \midrule

            Xiao et al. & Change Detection in 3D Point Clouds Acquired by a Mobile Mapping System \cite{ChangeDetection} & Ray tracing  & $\times$ & Post-mapping  \\\\
            
            Shishir Pagad et al. & Robust Method for Removing Dynamic Objects from Point Clouds \cite{Robust} & Ray tracing  & $\times$ & Post-mapping  \\\\
            
            Johannes Schauer et al. & The Peopleremover Removing Dynamic Objects From 3-D Point Cloud Data by Traversing a Voxel Occupancy Grid \cite{Peopleremover} & Ray tracing & $\times$ & Post-mapping \\\\
            
            Francois Pomerlea et al. & Long-term 3D map maintenance in dynamic environments \cite{pomerleau2014} & Visibility based  & $\times$ & Post-mapping  \\\\
            
            R. Ambruş et al. & Meta-rooms:
            Building and maintaining long term spatial models in a dynamic
            world \cite{metarooms}& Visibility based  & $\times$ & Post mapping \\\\

            Giseop Kim et al. & Remove, then Revert: Static Point cloud Map Construction using Multiresolution Range Images \cite{randr2020} & Visibility based  & $\times$ & Post-mapping  \\\\

            Hyungtae Lim et al. & ERASOR: Egocentric Ratio of Pseudo
            Occupancy-based Dynamic Object Removal for
            Static 3D Point Cloud Map Building \cite{erasor2021} & Segmentation & $\times$ & Post-mapping \\\\

            Xixun Wang et al. & Navigation of Mobile Robots in Dynamic Environments Using a Point Cloud Map \cite{indoor} & Segmentation &  $\checkmark$ & Post-mapping \\\\

            Schmid et al. & Dynablox: Real-Time Detection of Diverse Dynamic Objects in Complex Environments \cite{Dynablox} & Segmentation &  $\checkmark$ & Real-Time \\\\

            \bottomrule
        \end{tabular}
    \end{adjustbox}                      
    \label{tab:related_work_comparison}
    \par\medskip % This adds a little space before the note
\end{table*}

\subsection{Segmentation}

Segmentation strategies for removing dynamic points from point clouds are designed to distinguish between stationary and dynamic elements based on specific features of these points without using any ML based strategies. One notable method in this domain is Hyungtae Lim et al.'s ERASOR \cite{erasor2021}. Specifically tailored for urban settings, this technique operates on the premise that dynamic objects typically come into contact with the ground. By identifying height discrepancies between the original map and the current frame, ERASOR pinpoints dynamic elements. It then employs the Region-wise Ground Plane Fitting technique to efficiently separate static from dynamic points within these identified zones.

In a complementary approach, the Dynablox \cite{Dynablox} methodology offers an innovative way to detect moving objects through an online mapping-based strategy. This technique constructs and continuously updates a high-confidence map of clear space using the Voxblox framework. It identifies and labels areas completely free of objects. Dynamic objects are subsequently detected when they occupy these previously empty spaces, enabling real-time updates that reflect changes in the environment.

\subsection{Research Gap}

Ray-tracing techniques, while powerful, require significant computational resources and often struggle with incidence angle ambiguity. Conversely, methods based on visibility face difficulties with irremovable dynamic points. Table \ref{tab:related_work_comparison} provides a comprehensive comparison of the aforementioned methods. Notably, most dynamic object removal techniques have been applied in outdoor settings, to the best of our knowledge, with the exception of \cite{indoor}, which has been used indoors to create a 2D-occupancy map rather than a dynamic-free point cloud, and Dynablox \cite{Dynablox}. This study focuses on indoor environments characterized by slower movement dynamics compared to outdoor settings. It aims to detect dynamic points without needing pre-trained data and addresses incidence angle ambiguity in ray tracing and irremovable dynamic point challenges present in visibility-based methods. Moreover, the majority of dynamic point removal techniques utilize consecutive scans to detect and remove dynamic points and are unable to remove dynamic points present in a single map. The proposed approach, DynaHull, which focuses on density calculation, extends beyond the previously mentioned challenges. It operates effectively without needing a prior scans and tackles the issues of incidence angle ambiguity and irremovable dynamic points by assessing the dynamic features of each point in relation to its neighbors. Furthermore, occlusion becomes less severe since the density of the point is not related to the viewing angle. By capitalizing on these strengths, the proposed method holds promise for robust and efficient dynamic point removal in indoor environments.

\section{Methodology}\label{sec3}

This section is divided into three parts: "Equipment and Mapping", "DynaHull Algorithm", and "Validation". The first part offers details about the testing robot and the testing environment. The subsequent sub-section explains the DynaHull algorithm and its advantages. Lastly, "Validation" compares our method to other state-of-the-art methods.

\subsection{Equipment and Mapping}

\begin{figure}[h]
    \centering
    \begin{subfigure}{0.48\textwidth}
        \centering
        \includegraphics[width=\linewidth]{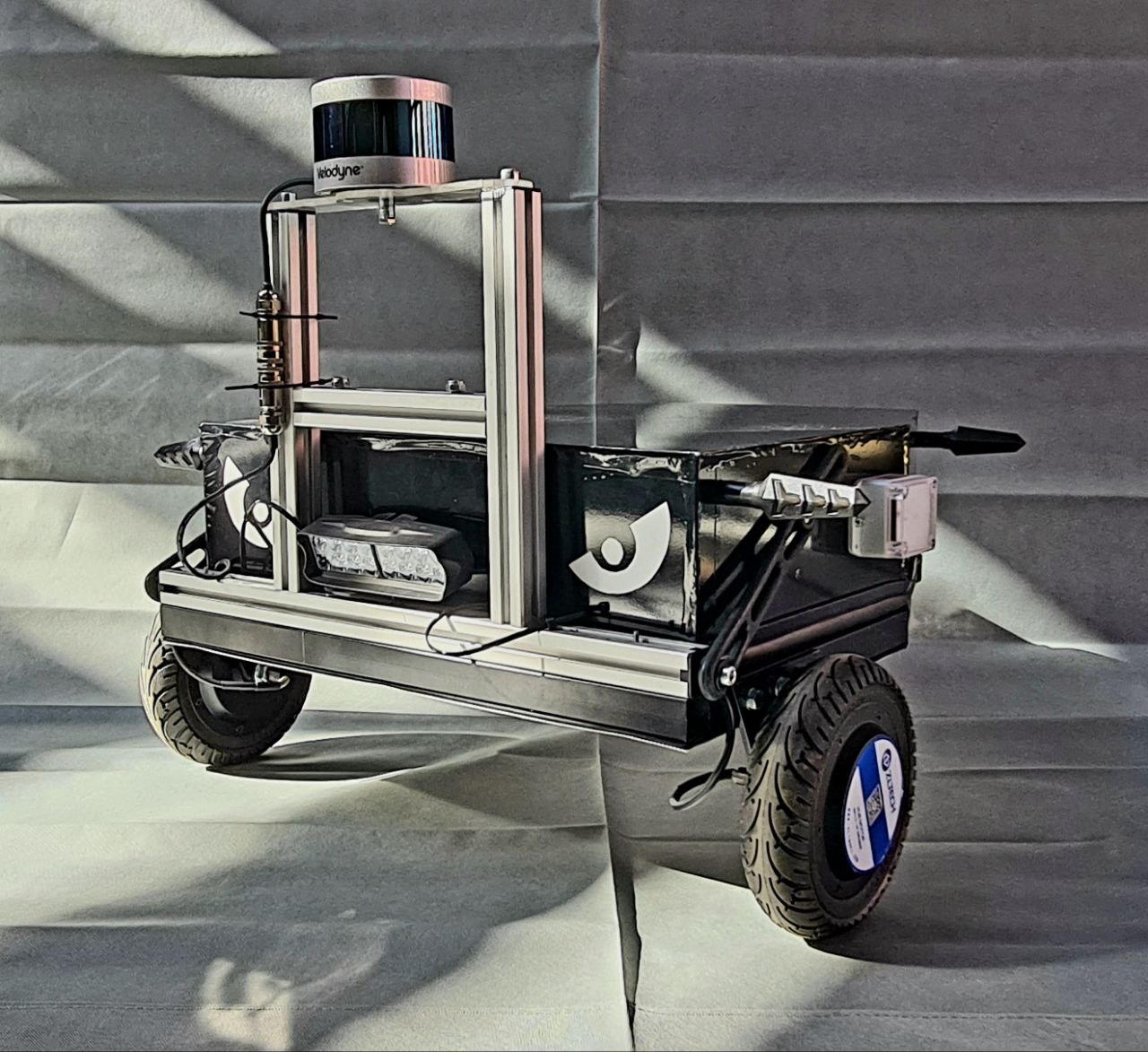}
        \caption{Dbot from side}
        \label{fig:Dbot1}
    \end{subfigure}\hfill
    \begin{subfigure}{0.44\textwidth}
        \centering
        \includegraphics[width=\linewidth]{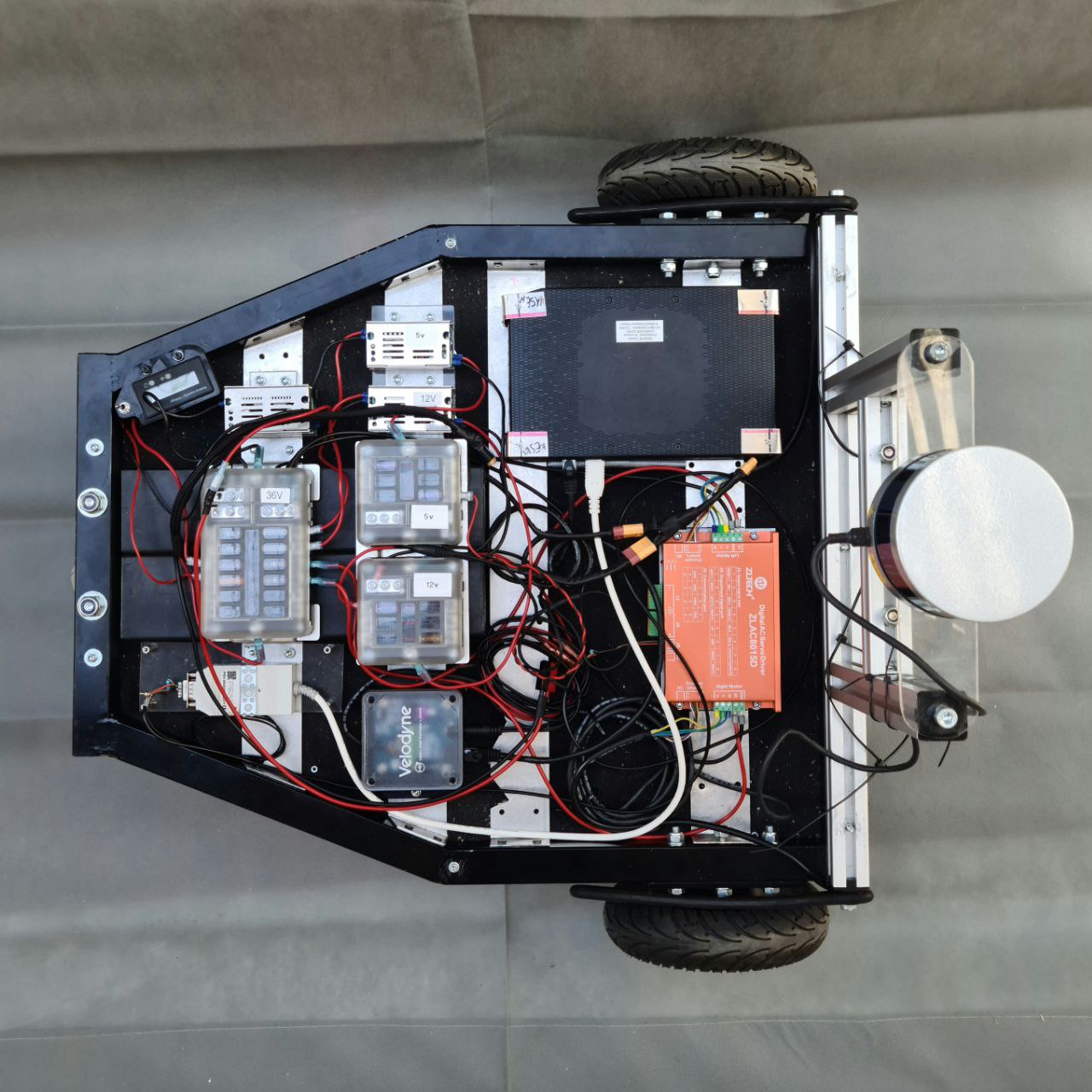}
        \caption{Dbot from top}
        \label{fig:Dbot2}
    \end{subfigure}
    
    \caption{Dbot- The robot used for collecting the Data}
    \label{fig:Dbot}
\end{figure}

The map containing dynamic points was collected using a differential wheel robot named DBot. DBot is equipped with tools, notably the Velodyne VLP 16 sensor, which captures 3D spatial points. The sensor has a 100-meter range and a 30-degree field of view. For optimal coverage, the sensor is mounted at the top of the DBot, as illustrated in Fig \ref{fig:Dbot}. The mapping in this study is accomplished by using an HDL graph slam package provided by Koide  \cite{hdl} in ROS (Robotic Operating System).

In this study, the map was created in the form of point cloud data (PCD) in an indoor environment at a university covering an area of approximately 750 square meters during a demo event in which about 100 people and numerous dynamic objects were present in the robots proximity. The maximum range of the Velodyne VLP16 was set to 75 meters and default parameters were used for HDL graph slam parameters. 
\begin{figure*}[h]
\centering
\includegraphics[width=1\linewidth]{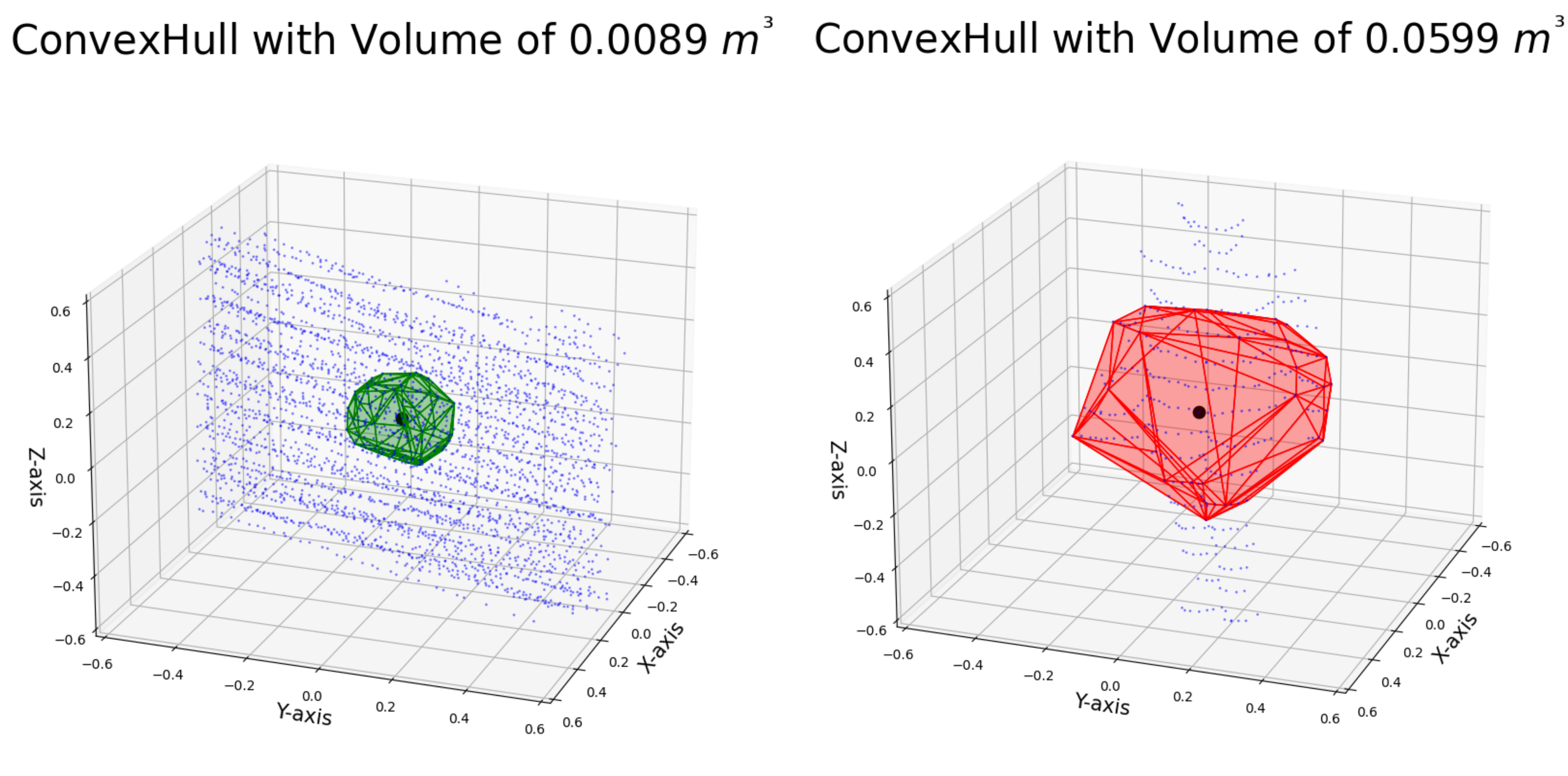}
\caption{Estimating convex hull volume between stationary and dynamic points: The left figure represents the stationary area (wall), while the picture on the right illustrates the dynamic area (human).}
\label{fig:ch}
\end{figure*}

\begin{figure}[h!]
\centering
\includegraphics[width=0.8\linewidth]{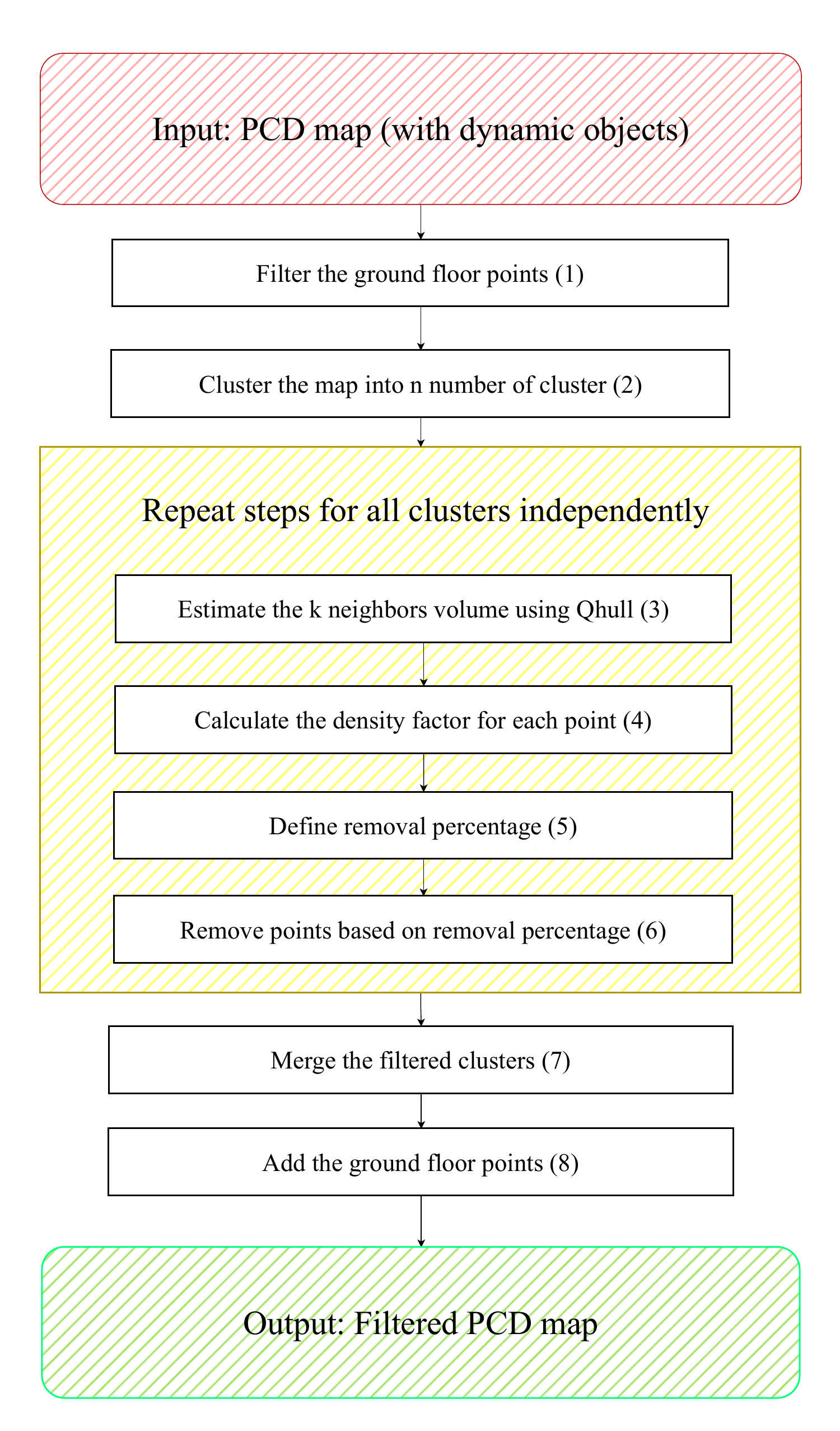}
\caption{Flowchart of DynaHull method}
\label{fig:Finalflowchart}
\end{figure}

\subsection{DynaHull}

The DynaHull method capitalizes on the observation that, after several scans, stationary points have a smaller convex hull volume compared to dynamic ones, as demonstrated in Fig \ref{fig:ch}.

On completing the mapping, the DynaHull process removes dynamic points as follows (the flowchart is illustrated in Fig. \ref{fig:Finalflowchart}): (1) DynaHull segments the ground points by estimating the slope of the ground points plane inspired by  \cite{groundfilter} and segments the floor points accordingly. (2) The map, free from ground points, is divided into $N_{clusters}$ clusters using Kmeans from \cite{sklearn}.
The number of clusters in this study is set to 5 as increasing the number of clusters did not have a noticeable change in performance of the filtering. (3) After clustering, nearest neighbor distances (in this study, $k=75$  number of nearest neighbors) of each point are calculated using sklearn's efficient KDtree \cite{sklearn}. These distances are then used to calculate the convex hull, which is the smallest convex set containing a given set of points. DynaHull integrates a divide-and-conquer algorithm known as Quickhull from Scipy \cite{scipy} to efficiently calculate the convex hull volumes $v$ (described in detail in the next paragraph). (4) After calculating the convex hull for each point, the point density factor is determined by $k/v$  (5) After assigning a density factor for each point, a removal percentage is defined based on the count of each cluster. Given a ground filtered point cloud, a number of clusters, and a range of removal [{min\_remove}, {max\_remove}] in this study "[5, 20]", the rescaling operation can be summarized as follows:

\begin{enumerate}
    \item For each cluster $i$ in $0, \ldots, N_{clusters} - 1$, count the points $N_i$.
    \item Let $N_{\min}$ and $N_{\max}$ be the minimum and maximum values among all $N_i$.
    \item For each $N_i$, rescale according to the formula:
\end{enumerate}

\begin{equation}
\label{eq:rescale}
R_i = \left( \frac{N_i - N_{\min}}{N_{\max} - N_{\min}} \right) \cdot (\text{max\_remove} - \text{min\_remove}) + \text{min\_remove}
\end{equation}
Where:

\begin{itemize}
    \item $R_i$ is the rescaled count (removal percentage) for each cluster $i$.
    \item $N_i$ is the original count for cluster $i$.
    \item $N_{\min}$ and $N_{\max}$ are the minimum and maximum counts across all clusters, respectively.
\end{itemize}

The resacling ensures that the clusters with low point counts will have a lower removal percentage, while the removal percentage increases as the cluster point count increases. This allows the algorithm to be flexible for areas with sparse stationary points which may be due to LiDAR expansion problems (also known as divergence of laser beams) or mapping limitations, such as obstacles or inaccessibility. (6). With the known $R_i$, Dynahull removes points in iterative steps by incrementally increasing a threshold value (starting with zero) based on the variances of all point's density factor until the required $R_i$ removal percentage in each cluster is achieved. These tasks are achieved by utilizing Open3D  \cite{open3d} and numpy \cite{numpy}. (7) This process is repeated for all clusters, and on completing the last cluster, these refined clusters are merged to reconstruct the map. (8) Finally, the ground floors are added back to the point cloud completing the DynaHull algorithm demonstrated in pseudo code in algorithm [\ref{alg:dynahull}]

\begin{algorithm}
\caption{DynaHull Algorithm}
\label{alg:dynahull}
\begin{algorithmic}[1]
\State Read \textit{point cloud data}
\State Remove the \textit{ground points} to get \textit{filtered\_point\_cloud} 
\State $\textit{labels} \gets \text{fit\_predict}(\text{KMeans}, \textit{filtered\_point\_cloud})$
\For{$i \gets 0$ \textbf{to} \textit{num\_clusters}}
    \State $\textit{clusters}[i] \gets \textit{filtered\_point\_cloud}[\textit{labels} = i]$
\EndFor
\State $\textit{cluster\_point\_counts} \gets |\textit{clusters}[i]|$ for each $i$

\State Scale point counts to $[\textit{min\_remove}, \textit{max\_remove}]$ removal percentage using the Equation [\ref{eq:rescale}] and save to re-scaled array
\For{each \textit{cluster} in \textit{clusters}}
    \State Calculate nearest distances using KDTree
    \State Calculate point density using QuickHull 
    \State Incrementally increase point density threshold till $[\textit{desired removal percentage} \gets \textit{re-scaled-array[cluster]}$
\EndFor
\State Concatenate filtered points to \textit{final\_cloud}

\end{algorithmic}
\end{algorithm}

\subsubsection{Quick hull}
The Quick hull in Step (4) operates as follows:  
\begin{itemize}
\item Initialization: The algorithm begins by identifying the extreme points on each axis (the smallest and largest) which are key components of the convex hull. Using these points, it forms an initial shape, often a tetrahedron. 
\item Face Processing: Each face of the initial shape is examined to find the point farthest from it, known as the ”furthest point.” 
\item Horizon Edge Search: For each face being examined, the algorithm searches for the ”horizon edge” at which a line from the furthest point intersects the shape. Faces visible from the furthest point are marked for removal. 
\item Constructing New Faces: Marked faces are removed, and new faces are created using the horizon edges as a base, linking the furthest point to these edges. 
\item Termination: This recursive process continues until no more points are found outside the convex hull. The algorithm concludes when no additional points can be seen from any face of the shape. The volume is returned by summing the volumes of the tetrahedrons. The volume of each tetrahedron is calculated using Equation \ref{eq:chv}, where \textbf{A}, \textbf{B}, \textbf{C}, and \textbf{D} represent the four vertices of the tetrahedron.
\begin{equation}
\label{eq:chv}
V_{tetrahedrons} = \frac{1}{6} \left| (\mathbf{A} - \mathbf{D}) \cdot ((\mathbf{B} - \mathbf{D}) \times (\mathbf{C} - \mathbf{D})) \right|
\end{equation}
\end{itemize}

\subsection{Validation}

Our methodology is evaluated using the DynamicMap Benchmark \cite{dynamicbench} by Kin-Zhang et al. consisting of well-known algorithms, such as ERASOR \cite{erasor2021}, Removert \cite{randr2020}, OctoMap + SOR \cite{octomap} implemented by Kin-Zhang and Dynablox \cite{Dynablox}. For ERASOR, the configuration for a semi-indoor environment from DynamicMap Benchmark was selected without any modifications. In Removert, the sequence field of the view parameter was adjusted from 50 to 29, aligning with the 30-degree FOV of the Velodyne VLP-16 used in our study. The OctoMap + SOR parameters remained unchanged same for Dynablox.

The core evaluation metric compares the map outputs of each method with a reference ground truth map created in a low-activity environment. Any ghost trails made by the robot operator were manually removed using the CloudCompare program. To compare generated maps to the ground truth, the distances of each point in the maps of each method are calculated to its closest point in the ground truth map. Afterwards, the mean absolute error (MAE), variance, root mean square error (RMSE) and 90th percentile error of these distance are recorded. Further comparisons employed the Chamfer Distance \cite{CD} and Earth Mover's Distance (EMD) \cite{POT} to assess the similarity between the methods and the ground truth map. The Chamfer Distance is calculated by computing the average of the squared distances from each point in one point cloud to its nearest point in the other point cloud, and vice versa, with a lower value indicating a closer match. EMD, also known as the Wasserstein distance, measures the minimum ’work’ required to transform one point cloud into another. In our study, the EMD is calculated using the ’emd2’ function from the Python Optimal Transport (POT) library, with maps uniformly downsampled to reduce computational requirements. 

Finally, this paper demonstrates the odometry performance of a map filtered by DynaHull and compares it with the odometry performance of a ground truth map and a dynamic map obtained by scan-matching live LiDAR data to each global map. Odometry data is collected using the ’HDL localization’ package provided by Koide  \cite{hdl}.

\section{Results}

\begin{figure}[h]
    \centering
        \centering
        \includegraphics[width=0.99\linewidth]{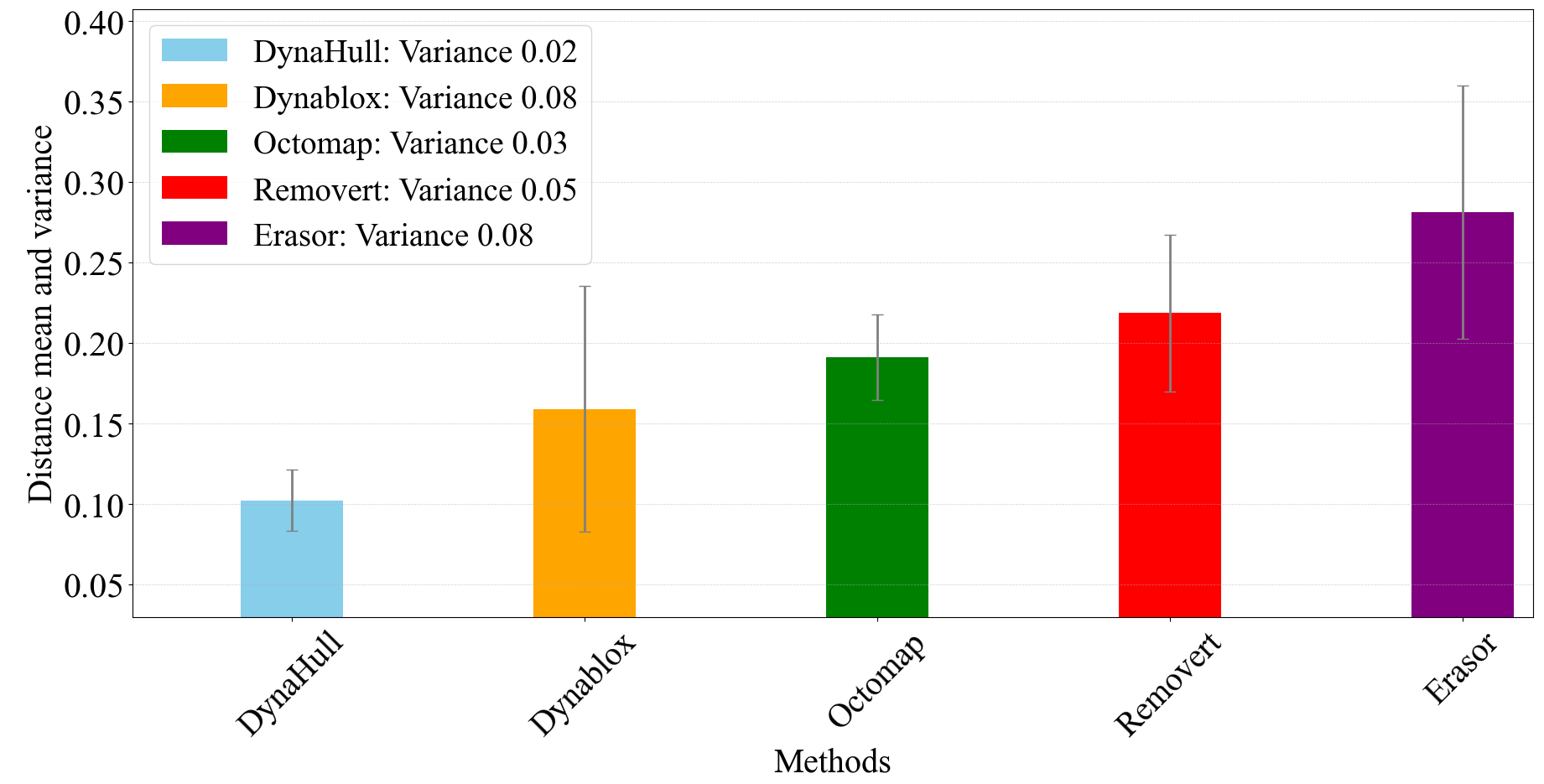}
        \caption{Mean and variance of distances between the points in each method compared to the ground truth}
        \label{fig:m&v}
\end{figure}

For each point in the map, the minimum distance to its corresponding point in the second map (ground truth map) is calculated. The means and variances of these distances are illustrated in Fig. \ref{fig:m&v}. It is evident that DynaHull has the lowest mean, followed by Dynablox, Octomap + SOR, Removert, and Erasor. In terms of variances, DynaHull remains the best, followed by Octomap + SOR, Removert, Dynablox, and Erasor. Table \ref{tab:mapscomparison} provides a further comparison between the filtered map and ground truth using five metrics: MAE, RMSE, 90th percentile error, CD, and EMD. It demonstrates that the similarity of the DynaHull filtered map with the ground truth is higher than that of other methods, particularly in terms of Earth Mover’s Distance. In general, DynaHull outperformed other methods in all metrics, while Dynablox had the second-best score in all metrics, followed by the Octomap+SOR method.

\begin{figure*}[h]
    \centering
        \centering
        \includegraphics[width=1\linewidth]{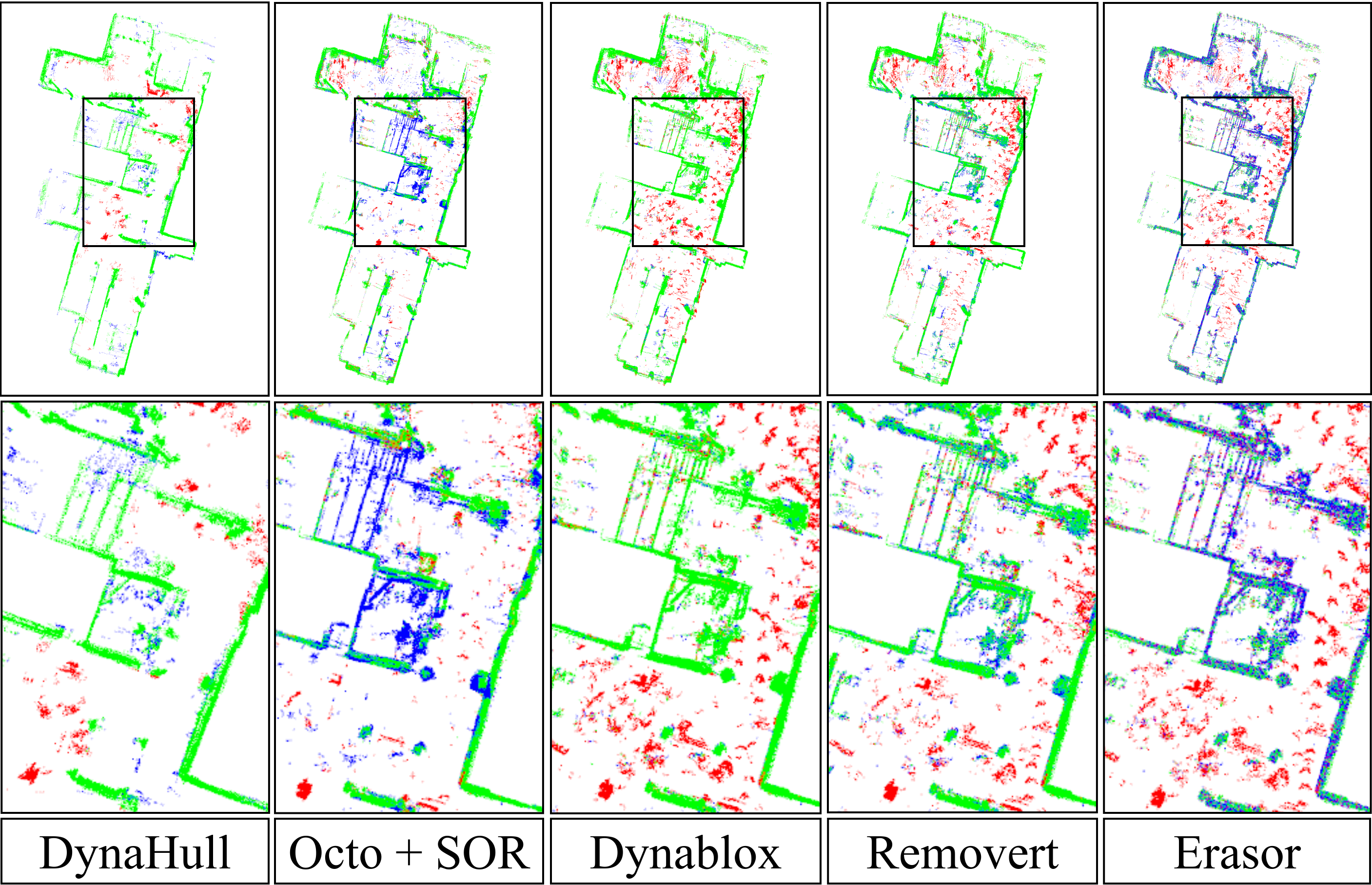}
        \caption{Comparison between DynaHull, OctoMap+SOR, Dynablox, Removert and ERASOR to the ground truth. The green points indicate true positives, the blue points represent false positives (incorrectly removed stationary points), and the red points are false negatives (dynamic points not removed).}
        \label{fig:mapcompare}
\end{figure*}

\begin{table*}[h]% Set font size to footnotesize for the entire table
\centering
\caption{Methods comparison between DynaHull and other methods}
\begin{tabular}{
    l
    *{9}{S[table-format=2.2]} % Adjusted to 9 columns to include the processing time
}

\toprule
{Method} & {MAE} & {RMSE} & {90 Percentile Errors} & {CD} & {EMD} \\
\midrule

DynaHull & \textbf{0.102} & \textbf{0.312} & \textbf{0.184} & \textbf{4.784} & \textbf{0.766} \\
Removert & 0.218 & 0.459 & 0.408 & 5.669 & 2.131 \\
Dynablox & 0.158&  0.398 &  0.287&  5.008&  0.790\\
OctoMap+SOR & 0.191 & 0.433 & 0.363 & 6.041 & 2.327 \\
ERASOR & 0.281 & 0.522 & 0.602 & 7.375 & 2.753 \\
\bottomrule
\label{tab:mapscomparison}
\end{tabular}
\end{table*}

To provide a visual comparison between all methods, Figure \ref{fig:mapcompare} shows true positives, false negative and false positive in green, blue and red. Note that the ground and ceiling points in each map have been removed for better visual comparison. On examining the Figure \ref{fig:mapcompare}, it is evident that other methods have excessively removed points from the walls which should be considered stationary, this is especially apparent in  ERASOR and removert. To clearly state the difference between each method the FP, FN and Processing time for each method is shown in Table \ref{tab:fnfp}. It is clear that DynaHull has the lowest FN and FP percentages. ERASOR performs exceptionally well in processing time but comes short in FP and FN with marginal difference. 

\begin{table}[h]
\centering
\caption{Comparison of FN, FP, Precision, Recall, and Processing Time }
\begin{tabular}{l S[table-format=1] S[table-format=1] S[table-format=1] S[table-format=1] S[table-format=1]}
\toprule
{Method} & {FP Percentage} & {FN Percentage} & {Precision} & {Recall} & {Processing Time (s)} \\
\midrule
DynaHull & \textbf{5.93} & \textbf{7.33} & \textbf{0.88} & \textbf{0.88} & 15.00 \\
Removert & 20.96 & 15.19 & 0.66 & 0.68 & 0.60 \\
Dynablox & 8.51 & 15.06 & 0.84 & 0.85 & 0.17 \\
OctoMap+SOR & 26.69 & 13.71 & 0.58 & 0.60 & 0.40 \\
ERASOR & 82.14 & 61.3 & 0.18 & 0.39 & \textbf{0.16} \\
\bottomrule
\label{tab:fnfp}
\end{tabular}
\end{table}

\begin{figure}[h]
    \centering
    \begin{subfigure}[b]{0.48\textwidth}
        \centering
        \includegraphics[width=\textwidth]{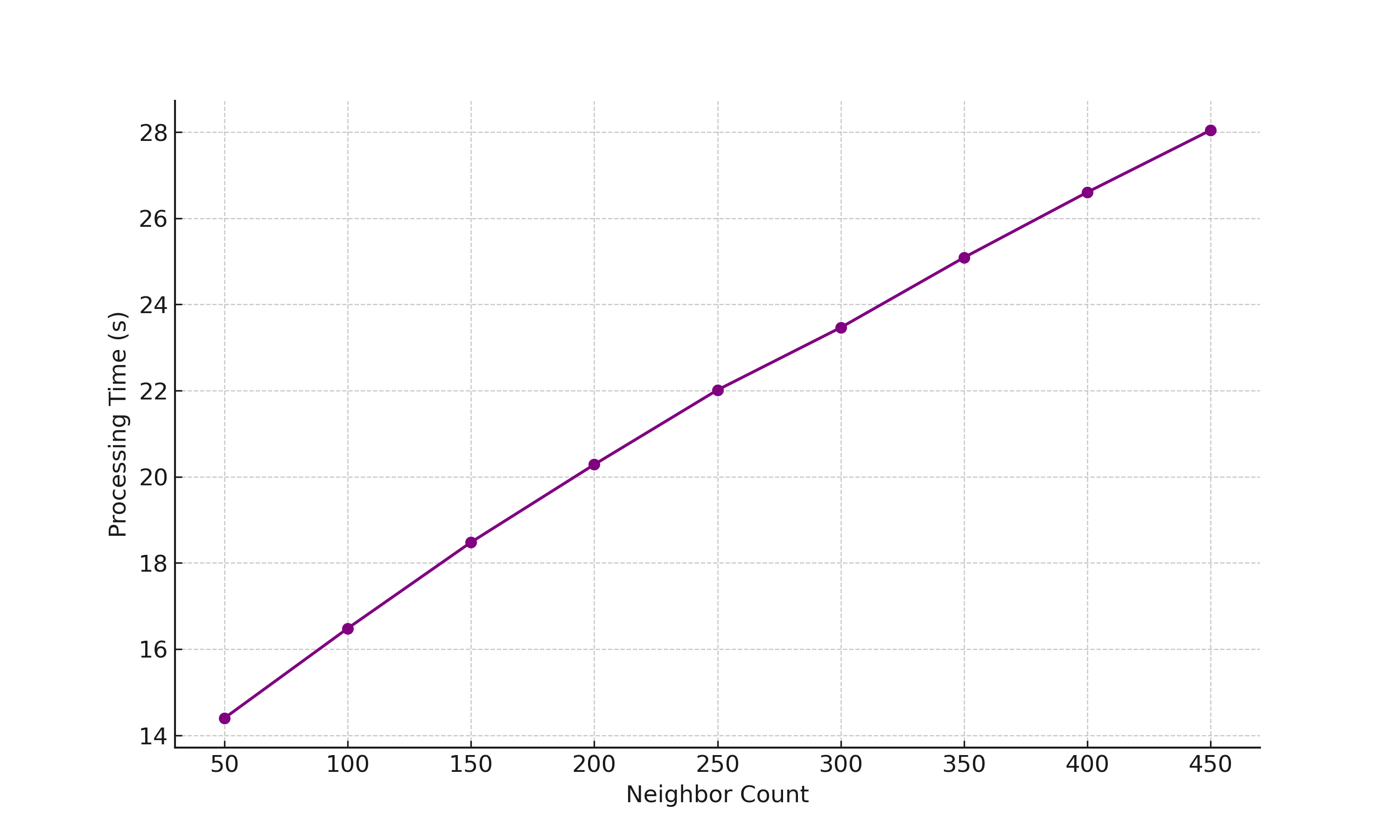}
        \caption{Processing time of DynaHull as the neighboring count increases}
        \label{fig:DynaHulltime}
    \end{subfigure}
    \hfill % Optional: add some horizontal spacing
    \begin{subfigure}[b]{0.48\textwidth}
        \centering
        \includegraphics[width=\textwidth]{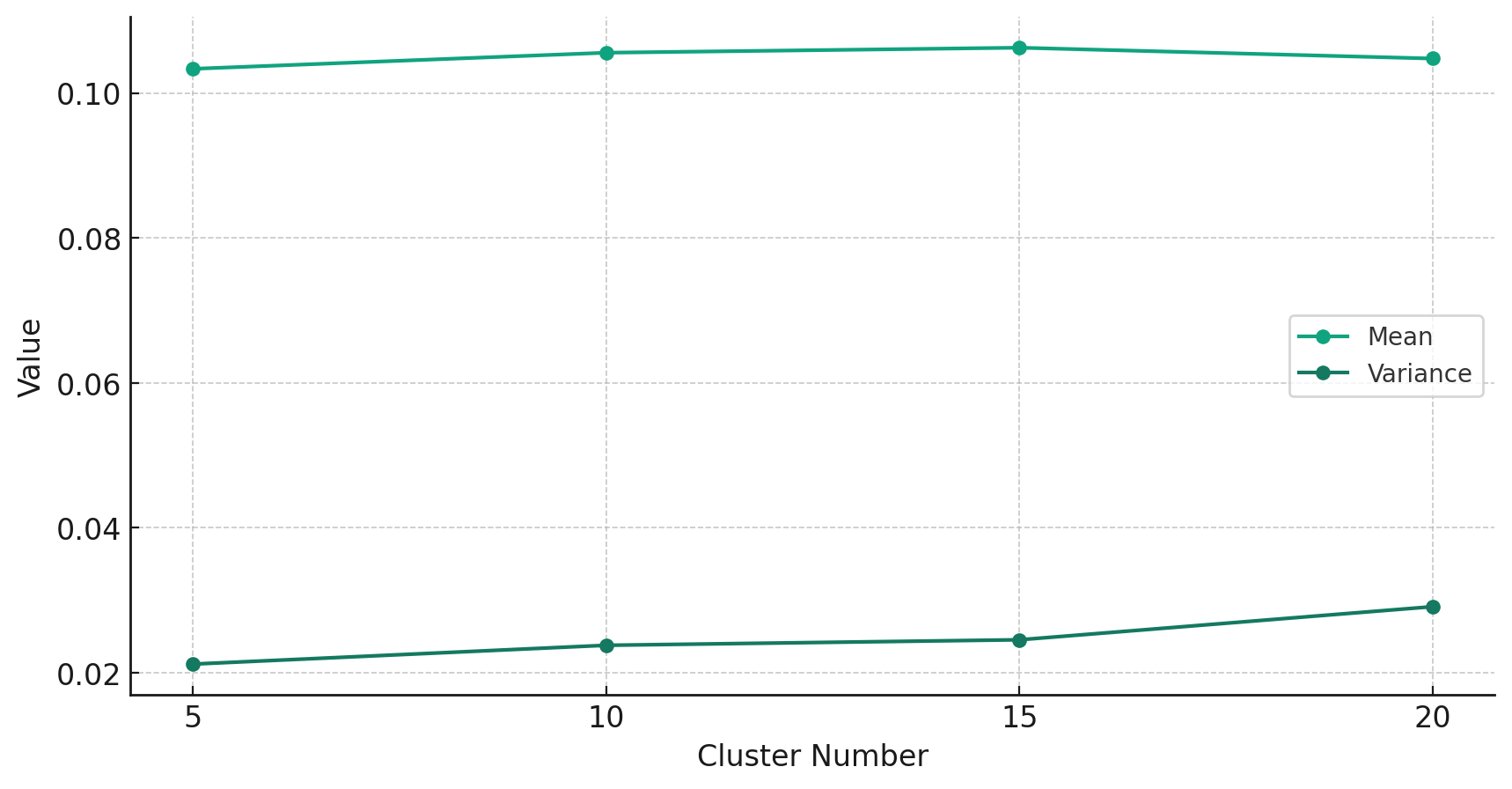}
        \caption{Cluster Count effect on mean and variance}
        \label{fig:clustercount}
    \end{subfigure}
    \caption{Comparison of computational aspects in different settings.}
\end{figure}

The two parameters affecting DynaHull's performance, namely neighboring point count and cluster count, are tested for their processing time and their respective effects on the mean and variances compared to the ground truth. It is observed that as the value of $k$ increases (as shown in Figure \ref{fig:DynaHulltime}), the processing time also increases linearly. Conversely, in Figure \ref{fig:clustercount}, as the number of clusters increases, the performance of DynaHull experiences negligible changes.

\begin{figure}
    \centering
        \includegraphics[width=1\linewidth]{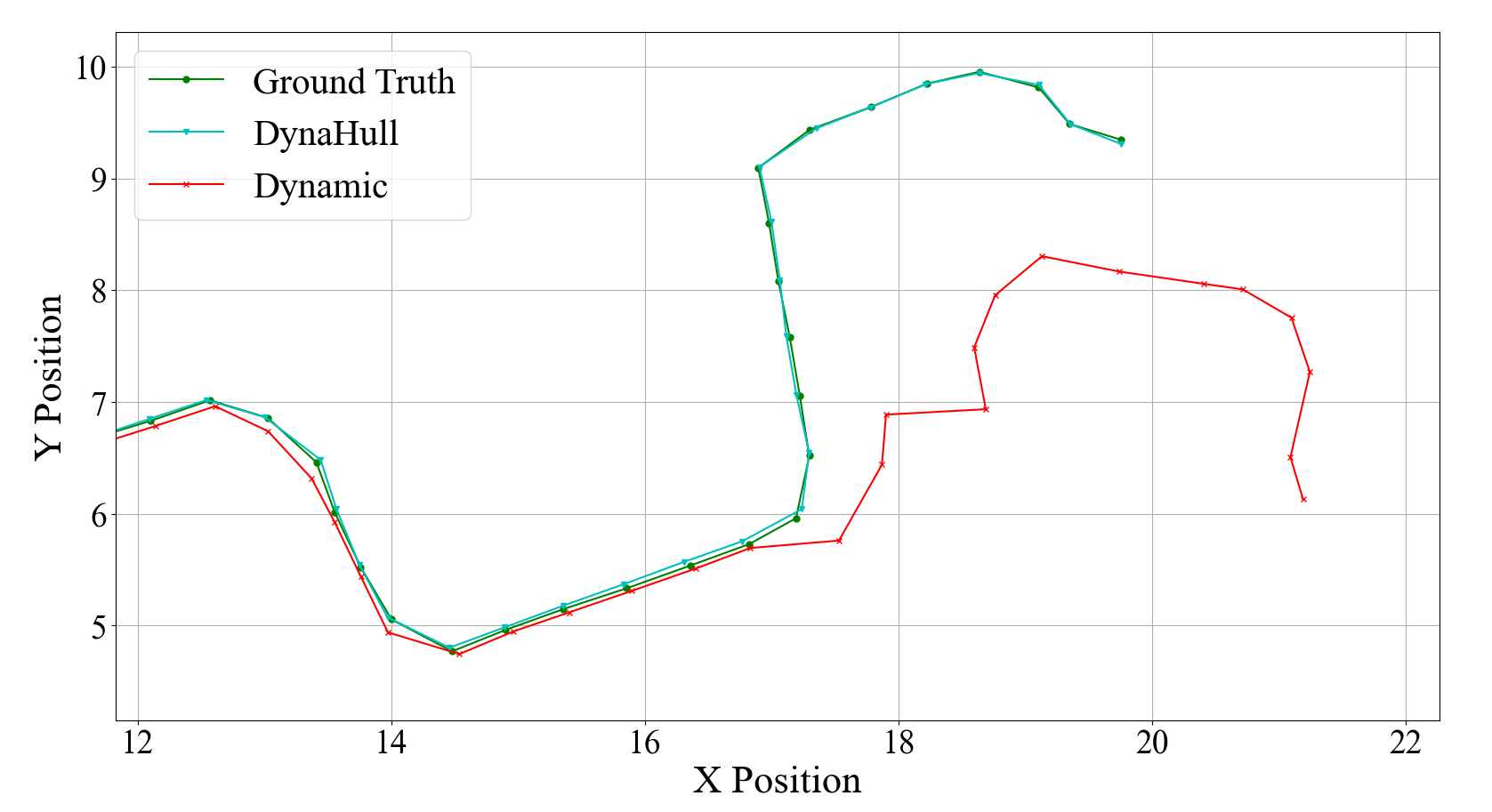}
        \caption{Odometry comparison between the ground truth, DynaHull, and Dynamic Map: The robot fails to localize itself after the third turn in dynamic map}
        \label{fig:fodom}
\end{figure}

To emphasize the importance of filtering and the effectiveness of DynaHull in localization, we compared the localization performance of a robot using a dynamic map, a DynaHull-filtered map, and a ground truth map. The corresponding odometry is presented in Fig. \ref{fig:fodom}. The figure illustrates that the DynaHull-filtered map excels in localizing similarly to the ground truth map, while the robot fails to localize itself in a dynamic map. The failure was particularly evident in turns; although the robot manages to localize itself during the first two turns in the dynamic map, it completely fails in subsequent turns. 

\section{Discussion}

The primary objective of this study was to improve the accuracy of indoors dynamic points filtering through density filtering. To reach this, a novel density-based method was developed. The proposed algorithm had a few parameters which affect the performance, mainly, the neighboring points count k and removal percentage. In terms of the neighboring points, a larger number might result in higher confidence of stationary points, however, an extensive increase of the k value results in more computational demand as shown in Figure \ref{fig:DynaHulltime}. Another factor influencing the performance of our method is the removal percentage. Increasing the removal percentage value enhances the confidence of dynamic point elimination, but an excessive value may lead to the removal of stationary points. In comparison Table \ref{tab:mapscomparison},  the DynaHull method outperformed other techniques in all metrics except for processing time. However, this post mapping algorithm can be further optimized to reduce the computational cost (For example running in GPU environment or in faster platform such as C++ with better parallelism and cache management) but the outcome will not be significant as this algorithm was designed for post mapping not real time.

Note that in indoor environments, other methods faced challenges in effectively detecting and removing dynamic points when objected to slow dynamism. Furthermore, while other methods utilized 32- or 64-channel LiDARs, our study employed a 16-channel LiDAR. This could explain the reason for these outdoor environment methods under-performing. This distinction is noteworthy as a higher point count is advantageous. This is notably more visible with the ERASOR method as the algorithm was built on the premise of the dynamic points being in contact with the ground. However, since the 16-channel LiDAR does not provide a sufficient number of ground points with which the algorithm can detect the dynamic points, ERASOR performs poorly in this scenario. For Dynahull, theoretically, increasing the number of LiDAR channels would enhance the distinction between stationary and dynamic points in the convex hull. However, this would also result in increased computational time due to the larger number of points to process. Considering the primary focus of our algorithm on indoor environments, the practicality of using 32 or 64 channel LiDARs may be questionable.

The other contributing factor is the low height of the Velodyne installed on the DBot compared to those installed on top of the autonomous vehicle. Elevating the 3D LiDAR results in better converge and can improve the performance of the benchmark methods. 

 One main drawback that affected the filtering performance in all maps was the inaccurate registration of points. Since the mapping was implemented in a highly dynamic environment, the points were not accurately registered due to the presence of a noticeable number of outliers (humans and other dynamic objects). This has caused walls or other stationary features to appear as artificially thickened due to cumulative mapping errors. Improving registration techniques or alignments can enhance the DynaHull method, since thinner walls will decrease the value of the convex hull volume estimates, thereby increasing the density of the stationary points.

\section{Summary and Conclusions}
%%\label{}

This study introduced the DynaHull method for removing dynamic points from point clouds post-mapping. The primary objective was to address the limitations of state-of-the-art techniques, including ray tracing, visibility, and segmentation methods. The results of our study demonstrate the better performance of our methods compared to other state-of-the-art techniques in all the presented metrics, namely, MAE, RMSE, 90th percentile error, CD, EMD, FP, FN except for processing time. The suggested future research directions include applying the DynaHull method across a broader array of scenarios and extending the range of hyperparameter optimization, especially for defining cluster count . This approach should incorporate more sophisticated algorithm designs, such as wall detection and machine learning-based segmentation of dynamic objects like humans. Additionally, the development of improved registration techniques tailored for highly dynamic environments is recommended to address the challenges in mapping registration.

\section*{Declarations}

\bmhead{Consent for Publication}
\hfill \break
Not applicable.
\bmhead{Funding}
\hfill \break
This project was funded by the HIP ROBOT project.
\bmhead{Authors' Contributions}
\hfill \break
Pejman Habibiroudkenar was responsible for developing the methodology, building the robot, collecting data, analyzing, and writing the report. Risto Ojala was behind the concept development, design of the research, setup, commenting on, and checking the final manuscript, as well as co-authoring the manuscript. Kari Tammi acquired funding, supervised the work, co-authored the manuscript, served as an advisor, and contributed to editing the final project.

%%\bmhead{Acknowledgments}

%%Acknowledgments are not compulsory. Where included they should be brief. Grant or contribution numbers may be acknowledged.

%%Please refer to Journal-level guidance for any specific requirements.

%\bibliographystyle{numeric}
\bibliography{sn-bibliography}% common bib file

\end{document}